%% file: camera-ready-sigconf.tex
\documentclass[sigconf]{acmart}

\copyrightyear{2024} 
\acmYear{2024} 
\setcopyright{acmlicensed}\acmConference[SA Conference Papers '24]{SIGGRAPH Asia 2024 Conference Papers}{December 3--6, 2024}{Tokyo, Japan}
\acmBooktitle{SIGGRAPH Asia 2024 Conference Papers (SA Conference Papers '24), December 3--6, 2024, Tokyo, Japan}
\acmDOI{10.1145/3680528.3687672}
\acmISBN{979-8-4007-1131-2/24/12}

\usepackage{booktabs} 

\usepackage[ruled]{algorithm2e} 
\usepackage{graphicx}
\usepackage[normalem]{ulem}

\SetAlFnt{\small}
\SetAlCapFnt{\small}
\SetAlCapNameFnt{\small}
\SetAlCapHSkip{0pt}

\begin{document}
\title{Frankenstein: Generating Semantic-Compositional 3D Scenes in One Tri-Plane}

\author{Han Yan}
\authornotemark[1]
\email{wolfball@sjtu.edu.cn}
\affiliation{
    \institution{Shanghai Jiao Tong University}
    \city{Shanghai}
    \country{China}
    \authornote{The contributions were made during their internships at Tencent XR Vision Labs.}
}

\author{Yang Li}
\email{liyang@mi.t.u-tokyo.ac.jp}
\affiliation{
    \institution{Tencent XR Vision Labs}
    \city{Shanghai}
    \country{China}
}

\author{Zhennan Wu}
\authornotemark[1]
\email{swwzn714@gmail.com}
\affiliation{
    \institution{The University of Tokyo}
    \city{Tokyo}
    \country{Japan}
}

\author{Shenzhou Chen}
\email{chenshenzhou@zju.edu.cn}
\affiliation{
    \institution{Tencent XR Vision Labs}
    \city{Shanghai}
    \country{China}
}

\author{Weixuan Sun}
\email{weixuansun7@outlook.com}
\affiliation{
    \institution{Tencent XR Vision Labs}
    \city{Shanghai}
    \country{China}
}

\author{Taizhang Shang}
\email{neoshang@tencent.com}
\affiliation{
    \institution{Tencent XR Vision Labs}
    \city{Shanghai}
    \country{China}
}

\author{Weizhe Liu}
\email{weizheliu1991@163.com}
\affiliation{
    \institution{Tencent XR Vision Labs}
    \city{Shanghai}
    \country{China}
}

\author{Tian Chen}
\email{tianchenucas@gmail.com}
\affiliation{
    \institution{Tencent XR Vision Labs}
    \city{Shanghai}
    \country{China}
}

\author{Xiaqiang Dai}
\email{xiaqiangdai@gmail.com}
\affiliation{
    \institution{Tencent XR Vision Labs}
    \city{Shanghai}
    \country{China}
}

\author{Chao Ma}
\authornotemark[2]
\email{chaoma@sjtu.edu.cn}
\affiliation{
    \institution{Shanghai Jiao Tong University}
    \city{Shanghai}
    \country{China}
    \authornote{Corresponding author.}
}

\author{Hongdong Li}
\email{hongdong.Li@gmail.com}
\affiliation{
    \institution{Australian National University}
    \city{Canberra}
     \country{Australia}
}

\author{Pan Ji}
\email{peterji1990@gmail.com}
\affiliation{
    \institution{Tencent XR Vision Labs}
    \city{Shanghai}
    \country{China}
}


\renewcommand\shortauthors{Yan et al.}

\begin{abstract}
{\bf Abstract:} We present Frankenstein, a diffusion-based framework that can generate semantic-compositional 3D scenes in a single pass. 
Unlike existing methods that output a single, unified 3D shape, Frankenstein simultaneously generates multiple separated shapes, each corresponding to a semantically meaningful part.
The 3D scene information is encoded in one single tri-plane tensor, from which multiple Signed Distance Function (SDF) fields can be decoded to represent the compositional shapes.
During training, an auto-encoder compresses tri-planes into a latent space, and then the denoising diffusion process is employed to approximate the distribution of the compositional scenes.
Frankenstein demonstrates promising results in generating room interiors as well as human avatars with automatically separated parts.
The generated scenes facilitate many downstream applications, such as part-wise re-texturing, object rearrangement in the room or avatar cloth re-targeting.
Our project page is available at:\url{https://wolfball.github.io/frankenstein/}.
\end{abstract}


%
%


%
%


\begin{teaserfigure}
\centering
\includegraphics[width=1.0\linewidth]{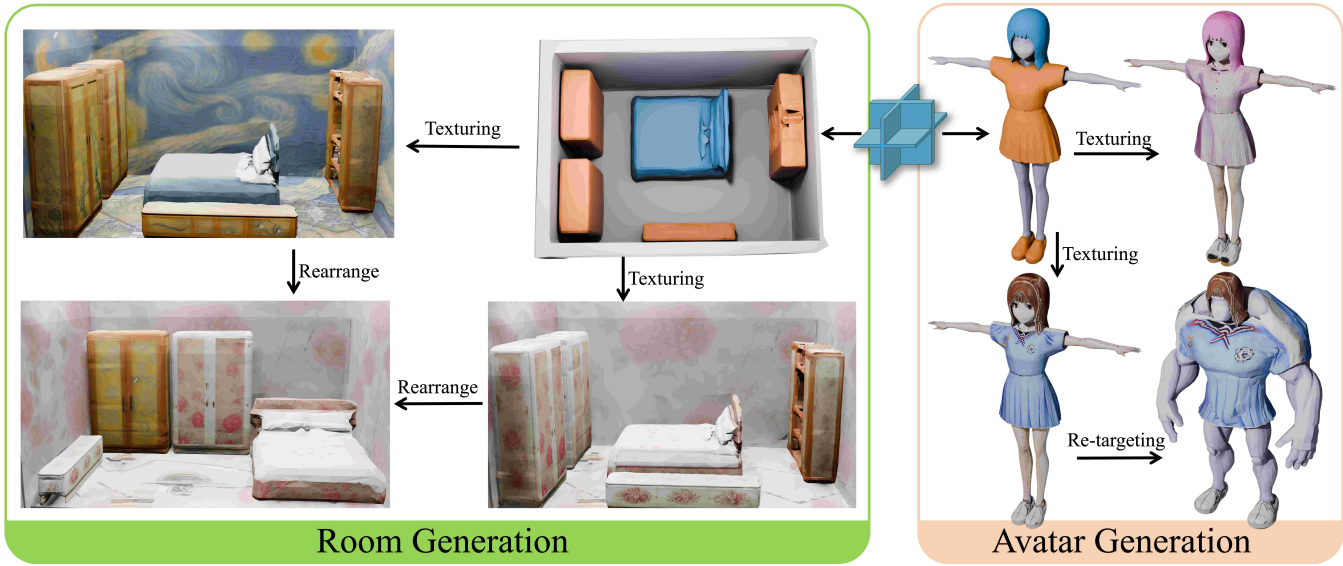}
\caption{
We present Frankenstein, a tri-plane diffusion-based framework that can generate semantic-compositional 3D scenes in a single forward pass, e.g., rooms (left) and avatars (right). 
The generated scenes enable customized controls, such as part-wise texturing, and room object rearrangement or avatar cloth re-targeting.
}
\label{fig:teaser}
\end{teaserfigure}

\maketitle

\input{samplebody-journals}

\end{document}

%% file: samplebody-journals.tex
\section{Introduction}
The creation of 3D assets with high-quality geometry is essential for many computer vision and graphics applications, including video gaming, film production, and AR/VR. 
%
3D generative models have achieved significant progress partly due to the adoption of the denoising diffusion models~\cite{sohl2015deep} and Transformers~\cite{vaswani2017attention_is_all_need}.
Recent works, such as~\cite{chou2023diffusion,qian2023magic123,gupta20233dgen,shi2023zero123++,gao2022get3d}, have demonstrated promising results on 3D assets generation using denoising diffusion.
These methods typically generate 3D data in the form of a single neural field, such as a Neural Radiance Field (NeRF)~\cite{nerf} or Signed Distance Field (SDF).  Consequently, semantic information contained in the generated 3D assets is entangled with other attributes in the representation.
%

%
However, downstream applications often require semantically-decomposed 3D shapes.
For example, in video games, a generated vehicle model should be able to be decomposed into a main body and its four roll-able wheels. Similarly, a 3D digital avatar shall be able to segment into parts like body, limbs, hair and apparel for realistic performance synthesis.
Although semantic 3D segmentation tools may be applied to segment a mesh into meaningful parts, so far the performance of existing methods is far from satisfactory, sometimes leading to incomplete fragments.
In this paper, we aim to develop a 3D generative AI model that enables directly generating semantic-compositional 3D scenes where each component has a complete shape.

Generating semantic-compositional 3D scenes poses two major challenges: 
Firstly, 1) a versatile 3D representation is required to jointly model the complete shape of multiple semantic components.
Secondly, 2) modeling the relationships between different semantic parts is complex. The relative positions between various parts should be semantically meaningful and physically plausible, e.g., avoiding interpenetration.

To address these challenges, we present~\textbf{Frankenstein}, a tri-plane diffusion-based approach that generates semantic-compositional 3D scenes.
The tri-plane is a tensor used to factorize the dense 3D volume
grid,
which is followed by an MLP to decode the 3D neural field signal.
We extend the tri-plane to represent compositional shapes by decoding multiple SDFs from a single tri-plane where each SDF contains the shape of a semantic class.
This representation allows simultaneous modeling of multiple complete shapes inside a single tri-plane tensor.
The training of Frankenstein consists of 3 stages:
First, 1) through per-scene fitting, we convert training scenes into tri-planes, which implicitly encode both the compositional shape information and the spatial relationships between components.
Then, 2) a variational auto-encoder (VAE) is trained to compress tri-planes into a latent tri-plane space which is significantly more compact and computationally efficient.
Lastly, 3) a denoising diffusion model is trained to approximate the distribution of the latent tri-planes.
We evaluate Frankenstein for two types of scenes: rooms composed of different classes of furniture; and avatars with clothing, hair, and body parts. 
Frankenstein generates meaningful compositional scenes, ensuring a clear distinction between scenes, and also showcases diversity in the shapes it generates.
The generated scenes facilitate downstream applications, including part-wise texturing, object rearrangement in the room, or avatar cloth re-targeting.

To summarize, our contributions are threefold:
\begin{itemize}
    \item We propose the first 3D diffusion model that can generate semantic compositional scenes in one tri-plane with a single forward pass. 
    \item We develop a robust coarse-to-fine optimization approach to produce high-fidelity semantic-compositional tri-planes. 
    \item We demonstrate the capabilities of our method for generating both room scenes and compositional avatars.    
\end{itemize}

\section{Related Work}
\subsection{3D Generation.} 
With the great success of diffusion model~\cite{sohl2015deep} in 2D domain~\cite{ho2020denoising,rombach2022high}, numerous studies have started investigating how to build 3D generation models.
There are mainly two technical solutions. 
One solution obtains 3D assets by distilling knowledge from pretrained 2D generator~\cite{poole2022dreamfusion,qian2023magic123}. 
DreamFusion~\cite{poole2022dreamfusion} proposed Score distillation Sampling (SDS) to optimize a NeRF model with images generated by a 2D generator.
Magic3D~\cite{qian2023magic123} utilized a coarse-to-fine two-stage strategy to improve both speed and quality. 
However, these methods usually suffer from the multi-view inconsistency problem and require time-consuming optimization.
The other technical solution involves directly training 3D generators using ground truth 3D data. 
Rodin~\cite{wang2023rodin} collected 100K 3D avatars using a professional synthetic engine and trained 3D-aware tri-plane diffusion model to generate NeRF~\cite{nerf} of the avatars.
Similar tri-plane diffusion-based methods can be found in~\cite{gupta20233dgen, nfd}.
LAS~\cite{zheng2023locally} instead directly performed diffusion on the 3D voxel grid.
LRM~\cite{hong2023lrm} formulated 3D generation as a deterministic 2D-to-3D reconstruction problem.
This paper builds upon the concept of tri-plane diffusion but introduces a major change: we encode multiple signed distance functions into a single tri-plane. This allows for the direct generation of compositional shapes within a single tri-plane denoising diffusion process.

\subsection{Compositional 3D Reconstruction and Generation.}
Research on reconstructing and generating holistic 3D scenes and objects has been widely investigated.
However, work on compositional scenes remains underdeveloped.
For 3D reconstruction, 
ObjectSDF~\cite{wu2022object} and ObjectSDF++~\cite{wu2023objectsdf++} introduced an object-composition neural implicit representation, which allows for the individual reconstruction of every piece of furniture in a room from image inputs;
DELTA~\cite{feng2023learning} introduced hybrid explicit-implicit 3D representations for the joint reconstruction of compositional avatars, such as the face and body, or hair and clothing, respectively.
In the field of 3D generation, AssetField~\cite{xiangli2023assetfield} proposed to learn a set of object-aware ground feature planes to represent the scene and various manipulations could be performed to rearrange the objects.
~\cite{po2023compositional,cohen2023set} jointly optimized multiple NeRFs, each for a distinct object, over semantic parts defined by text prompts and bounding boxes.
~\cite{epstein2024disentangled} and SceneWiz3D~\cite{zhang2023scenewiz3d} eliminated the requirements for user-defined bounding boxes by simultaneously learning the layouts.
Since the text could be problematically complicated when describing complex scenes, GraphDreamer~\cite{gao2024graphdreamer} used scene graphs as input instead.
For generating compositional avatars, some methods build upon SMPL~\cite{SMPL:2015} or SMPL-X~\cite{SMPL-X:2019} to facilitate layered avatar creation. Examples of such works include~\cite{hu2023humanliff, dong2024tela, wang2023disentangled, wang2023humancoser}.
In contrast to~\cite{po2023compositional,epstein2024disentangled}, our method generates compositional 3D scenes in a single reverse diffusion process rather than relying on multiple time-consuming Score Distillation Sampling (SDS) optimizations.
Moreover, our method is a general framework that does not rely on any shape prior, such as SMPL.

\begin{figure*}[!h]
  \centering
  \includegraphics[width=0.9\linewidth]{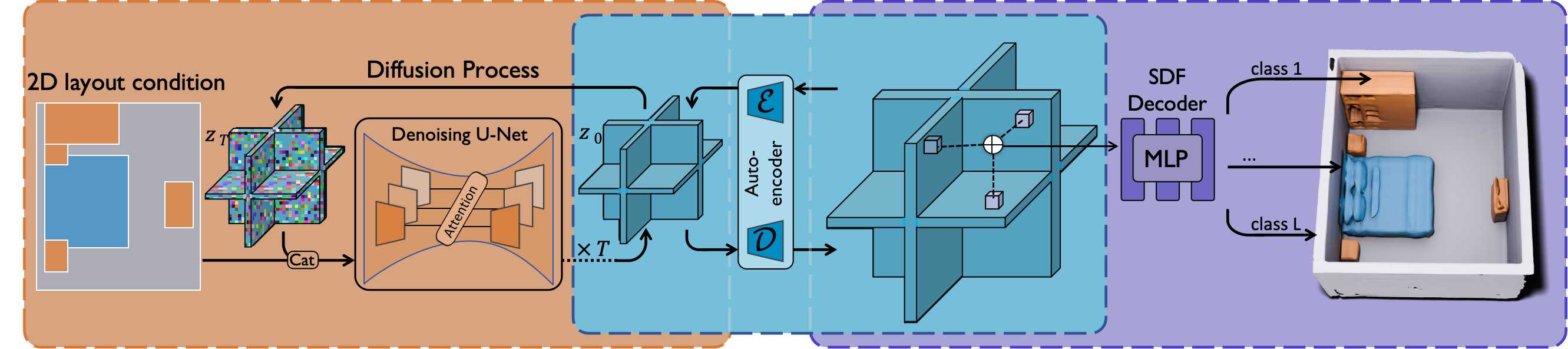}
  \caption{\textbf{Training pipeline of Frankenstein.} 
    \textcolor[RGB]{153,148,195}{ \textbf{ Tri-plane fitting:}} 
    training scenes are converted into tri-planes.
    %
    \textcolor[RGB]{103,168,192}{\textbf{ VAE training:}} 
    tri-planes are compressed into latent tri-planes via an auto-encoder.
    %
    \textcolor[RGB]{215,164,127}{\textbf{ Conditional denoising:}} 
    the distributions of latent tri-planes are approximated by a diffusion model conditioned on layout maps.
    During the inference process, given a 2D layout, the diffusion model denoises the noise to produce a latent tri-plane. This latent tri-plane is subsequently upsampled to a higher resolution by the VAE. Finally, a lightweight MLP decodes the high-resolution tri-plane into multiple semantic-wise SDFs.
    }
  \label{fig:ppl}
\end{figure*}

\subsection{Room-scale Scene Synthesis.}
Despite achieving impressive results in the synthesis of objects, the challenge of generating large and complex scenes persists~\cite{cogo2024survey,patil2024advances,chaudhuri2020learning}. 
The main difficulties lie in the high variance of scene geometry and complex positional relationships between different scene elements. 
Object-retrieval-based approaches initially generate the room layout based on deep convolutional network~\cite{wang2018deep,ritchie2019fast}, graph convolutional network~\cite{wang2019planit,zhou2019scenegraphnet}, recursive neural network~\cite{li2019grains}, generative adversarial networks (GAN)~\cite{nauata2020house}, auto-regressive transformer models~\cite{paschalidou2021atiss}, or diffusion models~\cite{tang2023diffuscene,zhai2023commonscenes,wen2023anyhome}, and then fill the layouts with objects that are searched from a given database.
However, these methods either do not consider the wall, an important element in the room scene, or will produce unnecessary penetration caused by inappropriate object placement.
Text2Room~\cite{hollein2023text2room} utilized a pre-trained 2D inpainting model to generate RGBD images and then fused these images into 3D. 
Ctrl-Room~\cite{fang2023ctrl} and ControlRoom3D~\cite{schult23controlroom3d} lifted generated images to 3D rooms with layout bounding boxes guidance.
The above 2D lifting-based approaches provide diverse contents but usually suffer from shape distortion.
CC3D~\cite{bahmani2023cc3d} used a conditional StyleGAN2 backbone to generate a 3D feature volume from a 2D semantic layout image and then used volume rendering to generate multi-view images of the room.
In contrast, our method simultaneously generates 3D indoor furniture, including walls, along with their arrangements, in a single process.


\section{Method}

This section provides details of our proposed framework (illustrated in Fig.~\ref{fig:ppl}) for the \textbf{room generation} task as an example.
Sec.~\ref{sec:stage1} introduces how we convert each 3D model into a tri-plane based implicit representation.
Sec.~\ref{sec:stage2} describes how we further encode tri-planes into a compact latent space.
Sec.~\ref{sec:stage3} presents how we realize controllable generation using diffusion models.

\subsection{Tri-Plane Fitting}\label{sec:stage1}

While both tri-plane~\cite{wang2023rodin,hong2023lrm} and volume grid~\cite{zheng2023locally,li2023diffusion} are commonly used as representations in 3D generation tasks, we opt for tri-plane because it is
1) more computationally efficient, and
2) more compatible with the advancements of image-based diffusion models.

Given a room $\mathcal{R}=\{\mathcal{R}_1,\dots,\mathcal{R}_L \}$ with $L$ classes of components, e.g., bed, cabinet, and wall, our goal is to train a tri-plane $\textbf{T}=\{\textbf{T}_{xy},\textbf{T}_{xz},\textbf{T}_{yz}\}\in\mathbb{R}^{3\times C\times R_h\times R_h}$ that can be decoded into $L$ separate SDFs by an MLP $\Phi$, and each SDF represents a class of room element.
The tri-plane $\textbf{T}$ is a spatial-aware tensor that can be queried by a 3D position $\textbf{p}\in\mathbb{R}^3$ and returns a feature vector $\textbf{f}\in\mathbb{R}^C$:
\begin{align}
    \textbf{f} &= Query(\textbf{T}, \textbf{p}),
\end{align}
where the $Query(\cdot)$ operation involves the summation of the 3 feature vectors that are bi-linearly sampled at $\textbf{p}$'s projection coordinates on three planes $\textbf{T}_{xy},\textbf{T}_{xz}$, and $\textbf{T}_{yz}$, respectively.

\begin{figure}[htb]
  \centering
  \includegraphics[width=0.9\linewidth]{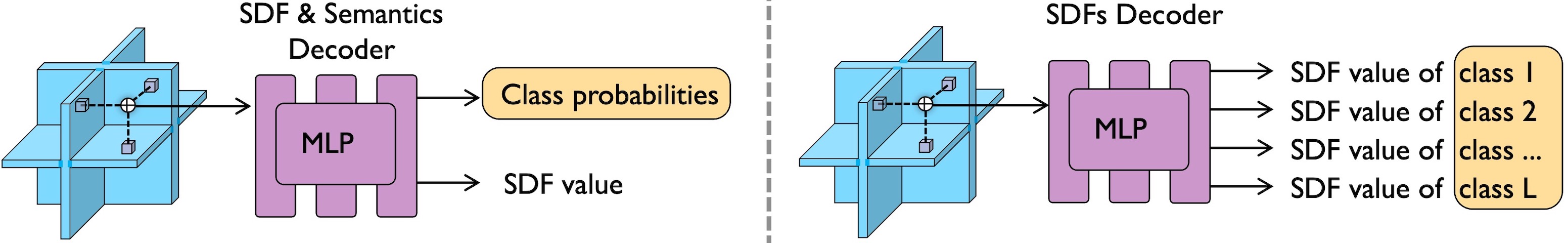}
  \caption{Two approaches to incorporate semantic information into neural fields. 
  }
  \label{fig:incor}
\end{figure}

\smallskip
\noindent
\textbf{Incorporating Semantics with Shapes.}
In neural field-based 3D reconstruction, numerous methodologies~\cite{zhi2021place,wu2022object,wu2023objectsdf++,yang2021learning} for incorporating semantic information with 3D shapes have been investigated.
Fig.~\ref{fig:incor} illustrates two ways, denoted as $\Upsilon$ and $\Phi$, from left to right. 
Specifically, $\Upsilon$ outputs a single SDF shape with a 3D semantic field
\begin{align}
 (d,\textbf{h})=\Upsilon(\textbf{f}),
\end{align}
where $\textbf{f}\in\mathbb{R}^C$ is the feature vector queried from the tri-plane, $d\in\mathbb{R}$ is the SDF value, $\textbf{h}\in\mathbb{R}^L$ is the one-hot vector indicating the class probability.
Then the scene can be decomposed by grouping the  points on the shape surface with their class labels. 
However this hard segmentation inevitably tear the surface of the shape and cause incompleteness in the decomposed parts.
In contrast,  $\Phi$ directly produces multiple shapes with each representing a distinct class. 
It reads
\begin{align}
(d_1,\dots,d_L)=\Phi(\textbf{f}),
\end{align}
where $d_i\in\mathbb{R}$ is the SDF value of the $i$-th class. We choose  $\Phi$ as our 3D representation  because it retains complete shapes for each class.  The output class-wise SDFs  is denoted as $\textbf{D} = (d_1,\dots,d_L)\in\mathbb{R}^L$.

\smallskip
\noindent
\textbf{Training Point Sampling Strategy.} \label{tpss}
The ground truth room scene is pre-processed in a compositional format: i.e., each part of the room, e.g., wall and chair, is represented by a separate triangle mesh.
We apply a semantic-aware on-surface point sampling strategy: 
given the mesh of the $l$-th class, 
we samples point set  $\mathbf{P}_l\in\mathbb{R}^{N_l\times 3}$ uniformly on the surface and compute the corresponding normal vectors as $\mathbf{n}_l\in\mathbb{R}^{N_l\times 3}$,
then we obtain $\{\mathbf{P}_{1},\dots,\mathbf{P}_L,\mathbf{n}_1,\dots,\mathbf{n}_L\}$ by merging samples from all classes.
We also sample off-surface points.
For SDF values, we randomly sample points $\mathbf{P}_{sdf}\in\mathbb{R}^{M\times 3}$ inside the  $[-1,1]^3$ cubic, and compute the class-wise SDF values as $\textbf{D}_{sdf}\in\mathbb{R}^{M\times L}$.
Additionally, we randomly sample point set $\mathbf{P}_{rnd}\in\mathbb{R}^{M\times 3}$ in $[-1,1]^3$ during every training epoch.

\smallskip
\noindent
\textbf{Class-Specific Geometry Loss.} 
To learn a unified tri-plane feature space, the weight of MLP $\Phi$ should be fixed for all training scenes.
Therefore, we jointly optimize $\Phi$ and the tri-planes of a subset of 10 rooms until convergence and then freeze $\Phi$ to perform per-scene fitting for the rest of rooms in the dataset.
The loss function is:
\begin{align}
    \mathcal{L}_{tri} &= \lambda_{1}\mathcal{L}_{eik} +\lambda_{2}\mathcal{L}_{sdf} + \lambda_{3}\mathcal{L}_{sur}+ \lambda_{4}\mathcal{L}_{nor}
\end{align}
The predicted SDF values $\{\tilde{\textbf{d}}_1\in\mathbb{R}^{N_1}, \dots, \tilde{\textbf{d}}_L\in\mathbb{R}^{N_L}, \tilde{\textbf{D}}_{sdf}\in\mathbb{R}^{M\times L}\}$ and normal vectors $\{\tilde{\mathbf{n}}_1\in\mathbb{R}^{N_1\times 3},\dots,\tilde{\mathbf{n}}_L \in\mathbb{R}^{N_L\times 3}, \tilde{\textbf{N}}_{rnd}\in\mathbb{R}^{M\times L\times 3}\}$ are obtained by applying $\Phi(\textbf{T}(\cdot))$ and $\nabla\Phi(\textbf{T}(\cdot))$ to all points in $\{\textbf{P}_{1}, \dots, \textbf{P}_L, \textbf{P}_{sdf}, \textbf{P}_{rnd}\}$, where the gradient $\nabla$ is performed using finite difference to approximate the normal vector.
Note that $\tilde{\textbf{d}}_{pcd}$ and $\tilde{\textbf{n}}_{pcd}$ remove the dimension for semantics, since the ground truth label is known for points in $\textbf{P}_{pcd}$.
Therefore the above terms are:
\begin{align*}
    \mathcal{L}_{eik} = \frac{1}{ML} (||\tilde{\mathbf{N}}_{rnd}||_2 - 1)^2, \quad \mathcal{L}_{sdf} = \frac{1}{ML}||\tilde{\mathbf{D}}_{sdf} - \mathbf{D}_{sdf}||_1,
\end{align*}
where $||\cdot||_1$ denotes L1-norm, and $||\cdot||_2$ denotes L2-norm.
The following terms are calculated in a class-specific way
\begin{align*}
    \mathcal{L}_{sur} = \sum_{l=1}^L \frac{1}{N_l} ||\tilde{\textbf{d}}_l||_1, \quad \mathcal{L}_{nor} = \frac{1}{L} \sum_{l=1}^L \frac{1}{N_l} ||\tilde{\textbf{n}}_l - \textbf{n}_l||_2.
\end{align*}
The above two terms alleviate the potential unbalanced point sampling, encouraging the reconstruction of semantic parts with small surface areas. 

\smallskip
\noindent
\textbf{Coarse-to-fine Optimization.} 
When calculating the surface normal vector using finite difference, discrete representations like tri-plane often face the gradient locality problem~\cite{li2023neuralangelo}.
Specifically, the gradient computation involves only neighboring features within a small step, which prevents it from affecting the optimization of tri-plane features beyond the adjacent grids. 
We found that this could cause noisy shape-fitting results and typically requires a large number of training points.
Neuralangelo~\cite{li2023neuralangelo} proposed to leverage the numerical gradient and progressive Level of Details (LoD) to address this problem.
However, this approach requires a pyramid of 3D feature grids, and generating such a pyramid data structure with diffusion model is not trivial.
In this paper, we fix the gradient locality problem with only one tri-plane.
Specifically, we develop a coarse-to-fine training strategy to fit a single high-resolution tri-plane.
First, we train a low-resolution tri-plane with $R_l^2$ to delineate the approximate semantics and shape of the scene.
Then, we upsample the tri-plane to $(2R_l)^2$ using bi-linear interpolation and refine the features at the larger tri-plane.
This process is repeated until the resolution reaches $R_h^2=2^\eta R_l$.
We discover that this straightforward strategy is effective in reconstructing compositional 3D shapes.

\begin{figure}[htb]
  \centering
  \includegraphics[width=0.75\linewidth]{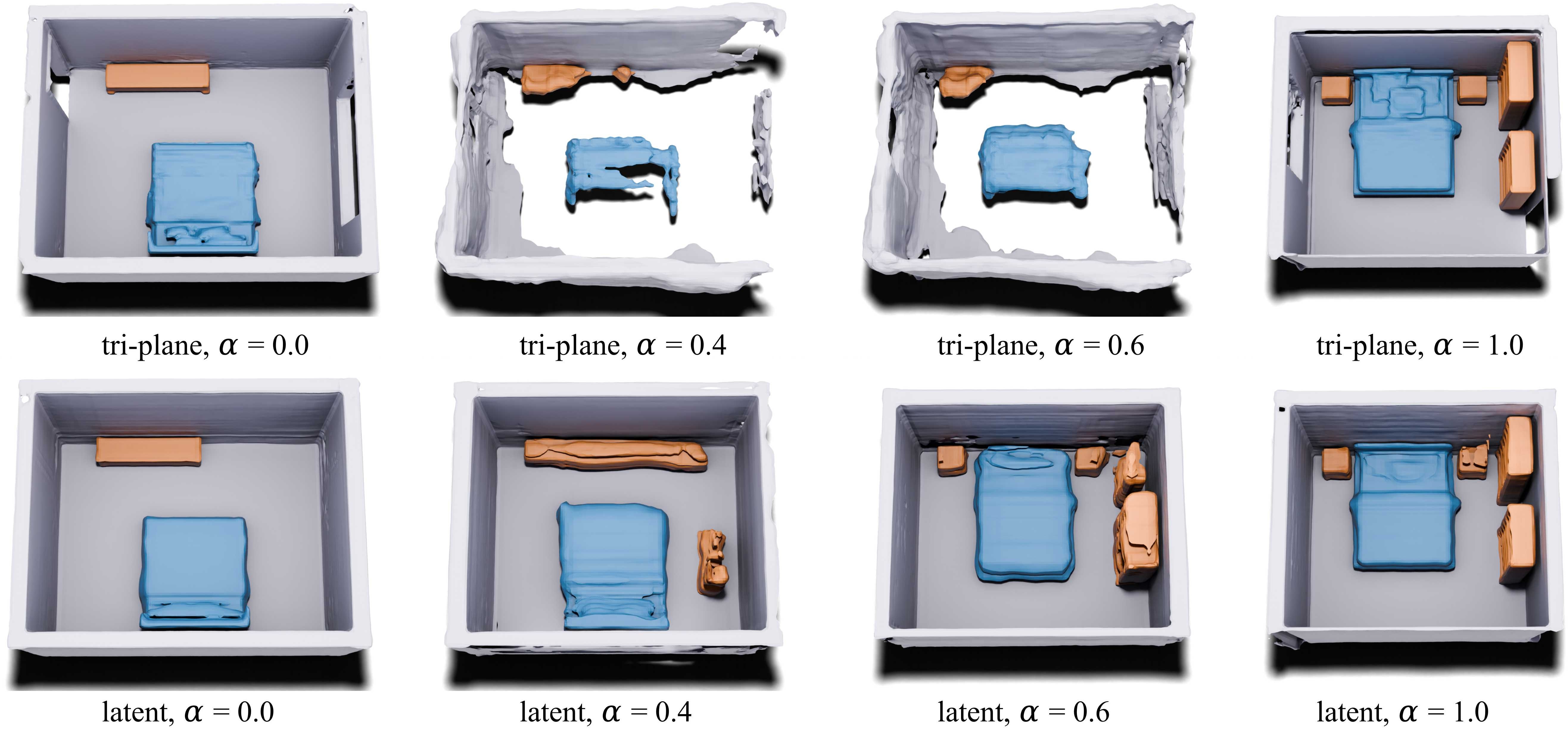}
\caption{Interpolation between two rooms on tri-plane space and latent tri-plane space. }
  \label{fig:interp}
\end{figure}

\subsection{VAE Training}\label{sec:stage2}
Directly training diffusion model on the fitted tri-plane $\mathbf{T}\in\mathbb R^{3\times C\times R_h\times R_h}$ is impracticable due to the high computational cost (the sizes are  $R_h=160$ and $C=32$ ).
Additionally, the tri-plane space does not form a continuous distribution. For example, the interpolation between two tri-planes appears meaningless (See Fig.~\ref{fig:interp}), which is not conducive to diffusion models.
To this end,  we follow BlockFusion~\cite{wu2024blockfusion} and employ an VAE to compress the tri-planes into a more compact and continuous latent space. The VAE performs the following mapping:
\begin{align}
    \mathcal E: \mathbb R^{3\times C\times R_h\times R_h}\mapsto\mathbb R^{3\times c\times r\times r},\quad \mathcal D: \mathbb R^{3\times c\times r\times r} \mapsto \mathbb R^{3\times C\times R_h\times R_h},
\end{align}
where $\mathcal E$ is the encoder, $\mathcal D$ is the decoder, and $r<R_h$ is the latent plane size.
VAE is trained with the following loss:
\begin{align}
    \mathcal{L}_{vae} &= \lambda_{rec}\mathcal{L}_{rec} + \lambda_{KL}\mathcal{L}_{KL} + \lambda_{tri}\mathcal{L}_{tri},
\end{align}
where $\mathcal{L}_{rec}$ is the L1-loss between reconstructed tri-plane $\tilde{\textbf{T}}=\mathcal{D}(\mathcal{E}(\textbf{T}))$ and the ground truth tri-plane $\textbf T$, 
$\mathcal{L}_{KL}$ is the KL-divergence.
Since the purpose is to learn a latent tri-plane that can faithfully represent the shape, we rely on $\mathcal{L}_{tri}$ as the dominate loss during VAE training.
 $\mathcal{L}_{tri}$ is computed similar as in Sec.~\ref{sec:stage1}, but using the queried features from the reconstructed tri-plane $\tilde{\textbf{T}}$.

\subsection{Conditional Denoising}\label{sec:stage3}
A well-trained VAE defines a latent space that is capable of not only reconstructing relatively complete room-scale geometry but also generating reasonable results through random sampling.
Conditional or unconditional diffusion model can be applied to this latent space for generation.
In this section, we take layout-conditioned room generation for example.

For a room $\mathcal{R}$ with $L$ classes,  the room layout is represented by the tensor $\mathcal{F}\in\mathbb{R}^{L\times r\times r}$, with each channel $\mathcal{F}_i\in\mathbb{R}^{r\times r}$ as a binary map indicating whether or not an object class is placed.  
$\mathcal{F}$ is computed by orthographically  projecting room elements onto the floor plane.
To condition on the floor layout, we follow~\cite{wu2024blockfusion} to directly concatenate the latent tri-plane and the floor layout as $z_0\in\mathbb{R}^{3\times(c+L)\times r\times r}$.
Since the $xz$ plane is aligned with the floor layout, we only concatenate $\mathcal{F}$ into the $xz$ plane of the latent tri-plane and pad the other planes with zeros:
\begin{align}
    z_0 = \{\hat{\textbf{T}}_{xy}\oplus\textbf{0}, \hat{\textbf{T}}_{xz}\oplus\mathcal{F}, \hat{\textbf{T}}_{yz}\oplus\textbf{0}\}.
\end{align}

In every iteration, we sample a timestep $t$ and add Gaussian noise $\epsilon$ to obtain $z_t$ from $z_0$:
\begin{align}
    z_t = \{\hat{\textbf{T}}_{xy}^{noised}\oplus\textbf{0}, \hat{\textbf{T}}_{xz}^{noised}\oplus\mathcal{F}, \hat{\textbf{T}}_{yz}^{noised}\oplus\textbf{0}\}.
\end{align}
And we train a U-Net shaped denoising backbone $\Psi$ to recover the latent tri-plane.
Instead of predicting the noise $\epsilon$ as in the original DDPM~\cite{ho2020denoising}, we predict $z_0$ in our task.
The loss function reads:
\begin{align}
    \mathcal{L}_{diff} = ||\Psi(z_t, \gamma(t)) - z_0||_2^2,
\end{align}
where $\gamma(\cdot)$ is a positional encoding function and $||\cdot||_2^2$ is MSE loss.

\smallskip
\noindent
\textbf{Dataset.}
Our room-scale scenes dataset is obtained from 3D-FRONT~\cite{fu20213d} and 3D-FUTURE~\cite{fu20213dfuture}.
The following pre-processes are performed:
1) The raw meshes are transformed into watertight ones using Blender's \textit{voxel remeshing} tool.
2) The positions of the furniture are adjusted if penetration is detected. This resolves the collision problem and provides a physically plausible training scene.
The final dataset contains 2558 bedrooms with 3 classes \{wall, bed, cabinet\}.
For VAE training, the fitted tri-planes are augmented by 8 times via rotations and flips.

\smallskip
\noindent
\textbf{Implementation Details.} 
The hyper-parameters are empirically set to $L=3, C=32, R_h=160, R_l=5, M=300000, c=4,r=40$.
For tri-plane fitting, we weigh the individual losses with $\lambda_1=0.2, \lambda_2=10.0, \lambda_3=10.0, \lambda_4=0.5$, and equally sample $N_1=N_2=N_3=100000$ points for each category.
For VAE training, we instead sample $N_1=20000, N_2=N_3=5000$ points and set $\lambda_{rec}=0.1, \lambda_{KL}=1.0, \lambda_{tri}=1.0$. 
Channel-wise normalization~\cite{nfd} is applied to the (latent) tri-plane features during VAE and diffusion training 
(Sec.~\ref{sec:ablation} shows ablation study of normalization strategy).
We run tri-plane fitting on multiple NVIDIA T10 GPUs, and the other stages on 8 NVIDIA V100 GPUs.
We assess the reconstruction quality using Chamfer Distance (CD).

\begin{figure*}[ht!]
\centering 
\includegraphics[width=0.8\linewidth]{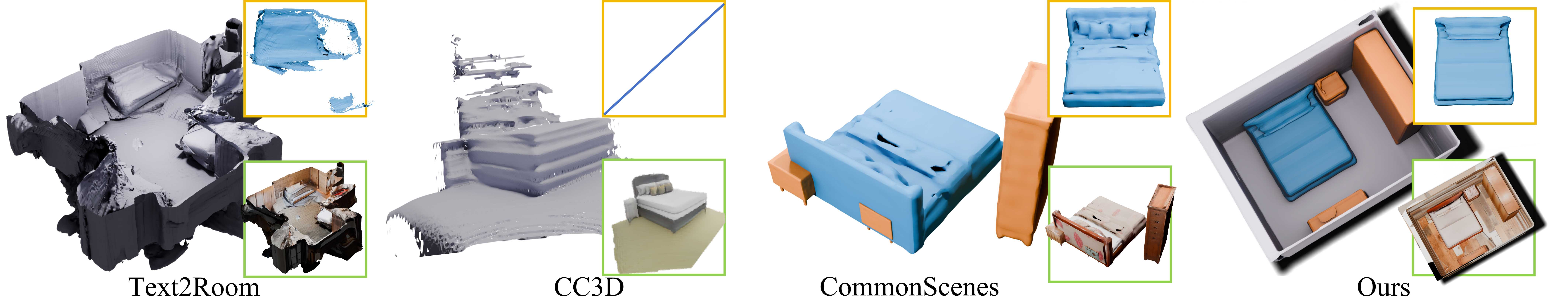}
\caption{ \textbf{Qualitative room generation results}. The prompt for Text2Room~\cite{hollein2023text2room} is ``a wooden style bedroom with a king-size bed and large wardrobes.'' The textures of CommonScenes'~\cite{zhai2023commonscenes} and ours are generated using SyncMVD~\cite{liu2023text} based on prompt ``wooden''. The geometry of CC3D~\cite{bahmani2023cc3d} is reconstructed from a point cloud extracted from depth images using ball-pivoting.}
  \label{fig:t2r}
\end{figure*}

\begin{table}
    \caption{
    \textbf{Quantitative room generation results.}}
    \label{tab:user-study}
    \centering
    \scalebox{0.8}{
        \begin{tabular}{l|ccc|c}
        \toprule
          & \multicolumn{3}{c|}{Whole Scene}  &  Component-Wise \\
          
        Method & RGQ$\uparrow$ & RLQ$\uparrow$& RLC$\uparrow$& CGQ$\uparrow$\\
        \midrule
        Text2Room~\cite{hollein2023text2room} & 2.18 & 2.18 & -& 1.05\\
        CC3D~\cite{bahmani2023cc3d} & 1.50 & 1.55 & 2.06 & -\\
        CommonScenes~\cite{zhai2023commonscenes} & 3.00 & 2.57 & -& \textbf{3.72}\\
        Frankenstein (Ours)   & \textbf{4.33} & \textbf{4.35} & \textbf{3.83}& 3.63\\
        \bottomrule
        \end{tabular}
    }
\end{table}

\begin{table}
    \caption{
    \textbf{Physical Plausibility Evaluation.} Physical plausibility refers to the rate of non-interpenetration among 50 scenes (test set).}
    \label{tab:penetration}
    \centering
    \scalebox{0.75}{
        \begin{tabular}{c|c|c}
        \toprule
         & CommonScenes~\cite{zhai2023commonscenes} & Frankenstein (Ours)\\
        \midrule
        Physical Plausibility(\%) $\uparrow$ & 40 & 84\\
        \bottomrule
        \end{tabular}
    }
\end{table}

\subsection{Compositional Room Generation.}
%
The evaluation considers two aspect: the quality of the whole scene and the quality of each component in a scene.
We compare with the following baselines:
1) Text2Room~\cite{hollein2023text2room}, a holistic room generation approach (as Text2Room does not generate separable meshes for each scene component, we obtain the separated object shapes by running the mesh semantic segmentation tool~\cite{Schult23ICRA}); 
2) CC3D~\cite{bahmani2023cc3d}, a layout conditioned room generation method (CC3D's component-wise results is omitted as running segmentation tool~\cite{Schult23ICRA} on CC3D fails);
and 3) CommonScenes~\cite{zhai2023commonscenes}, a scene-graph-based room generation baseline.
Fig.~\ref{fig:t2r} shows the qualitative room generation results.
Overall our model produces superior room geometry and more plausible layouts than Text2Room.
The segmented object's mesh from Text2Room tends to be broken.
Although CC3D also enables layout control, it often fails to extract valid geometry from arbitrary camera views due to the use of the NeRF representation, i.e. it is hard to obtain a complete mesh of room.
CommonScenes produces detailed shapes but struggles with object interpenetration and does not generate walls.
We conducted user studies for evaluation: over 60 participants were invited to rate the generation results using a score ranging from 1 to 5.
The user study includes the following metrics: Room Geometry Quality (RGQ), Room Layout Quality (RLQ), Room Layout Consistency (RLC), and Component-Wise Geometry Quality (CGQ). 
RLC measure the consistency between the input layout prompt and generated shape. 
%
Quantitative results are shown in Tab.~\ref{tab:user-study}.
Our method achieves the best whole scene scores (RGQ, RLQ, and RLC), and comparable per-object score (CGQ) to CommonScenes which generates each object separately and then combines them together.
However, CommonScenes may lead to object penetration, while our method produces physically more plausible results (c.f. Tab.\ref{tab:penetration} \& Fig.\ref{fig:t2r}).
Though there are several widely-adopted metrics to evaluate the generated layouts, such as the Fréchet inception distance (FID), we believe it is unfair as the training settings of the baselines are different from ours.

\begin{figure}[htb]
  \centering
    \includegraphics[width=0.8\linewidth]{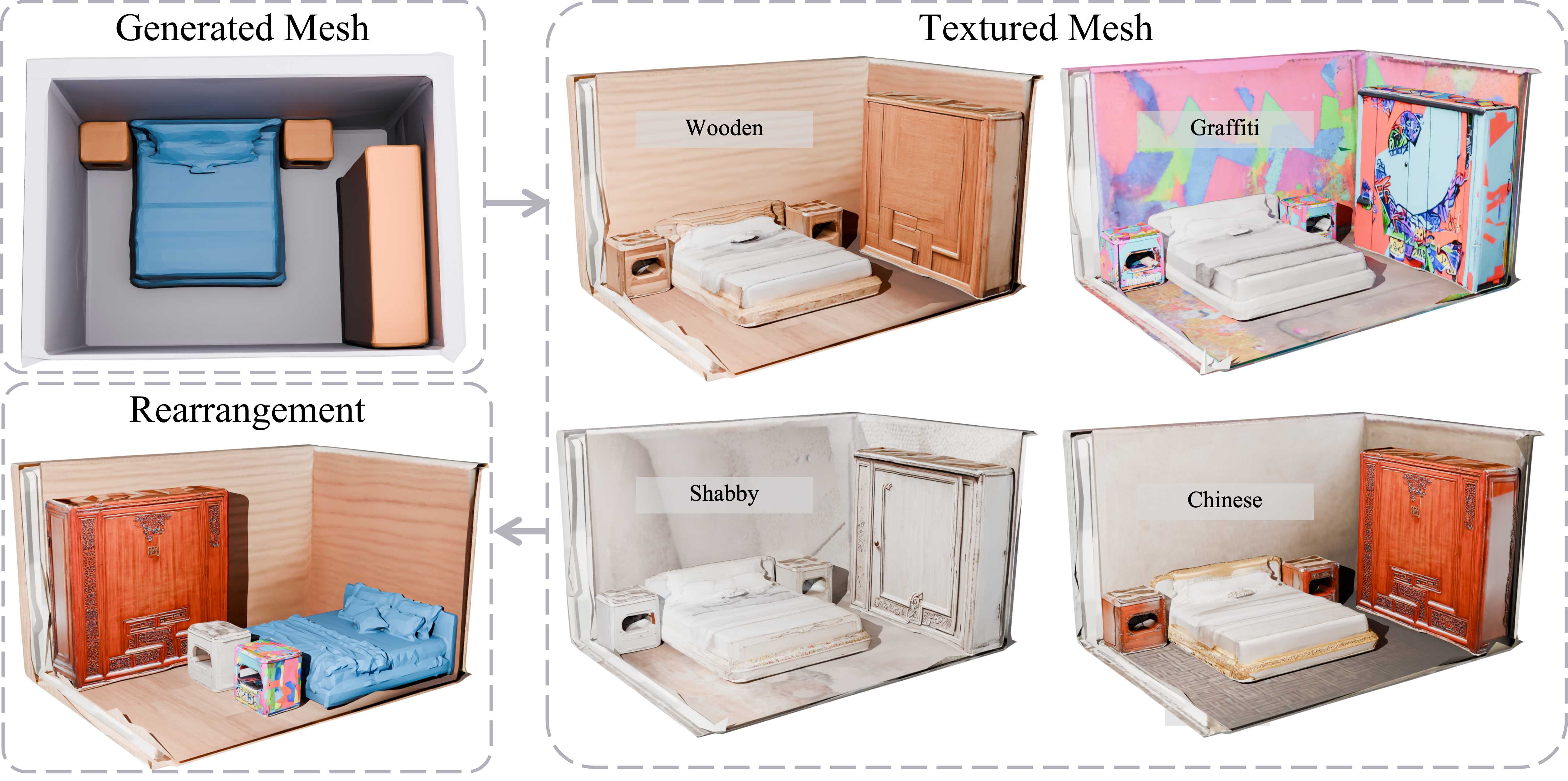}
    \caption{Applications of semantic-compositional room generation. Each component (\textcolor[RGB]{149,148,160}{wall}/\textcolor[RGB]{85,139,184}{bed}/\textcolor[RGB]{206,143,95}{cabinet}) is textured using ~\cite{chen2023text2tex} with \textit{wooden, graffiti, shabby and Chinese} styles. Rearrangement can also be applied.}
  \label{fig:refine}
\end{figure}

Fig.~\ref{fig:refine} displays various applications applied to our generated meshes. 
Semantic-compositional scenes provide semantic priors, which are compatible with off-the-shelf object-targeted texturing models. 
We utilize\cite{chen2023text2tex} to texture individual scene components in four styles.
Additionally, object rearrangements can be performed to customize the room layout and appearance. 
Moreover, retrieval and refinement can be implemented as a post-processing stage to further enhance the quality of 3D models.

\begin{figure*}[ht!]
  \centering
    \includegraphics[width=0.8\linewidth]{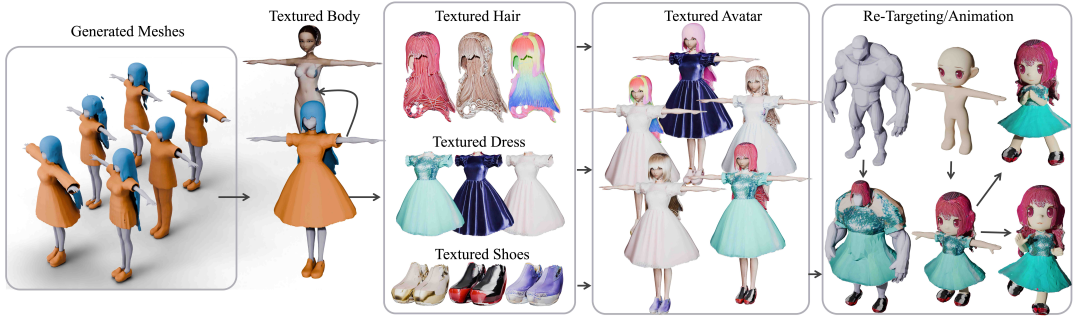}
    \caption{Applications of semantic-compositional avatar generation. Each component (\textcolor[RGB]{149,148,160}{body}/ \textcolor[RGB]{85,139,184}{hair}/ \textcolor[RGB]{206,143,95}{apparel}) is textured using ~\cite{chen2023text2tex,liu2023text} and assembled into different  avatars. Re-targeting and animation can also be applied.}
  \label{fig:lc}
\end{figure*}

\subsection{Compositional Avatar Generation}
In addition to room scenes, we also demonstrate the capacity of Frankenstein for compositional avatar generation.
In this experiment, Frankenstein is trained and tested on a dataset of around 10K cartoon avatars from Vroid~\cite{chen2023panic3d}, with each avatar being decomposed into body, clothing, and hair parts.
These avatar scenes pose a bigger challenge than the room scenes, as the level of shape entanglement between two parts of an avatar, e.g., hair and body, is much higher than that in room furniture and walls.
Nonetheless, Frankenstein still manages to successfully generate clean and separable shapes.
As shown in Fig.~\ref{fig:lc}, the generated compositional avatar facilitates numerous downstream applications, including component-wise texture generation, random cloth swapping, cloth re-targeting, and automatic rigging and animation.
The hair and clothing textures are generated using~\cite{chen2023text2tex}.
Cloth re-targeting is done by first registering the source and target bodies with a SMPLX~\cite{SMPL-X:2019} model and then transforming the cloth based on a dense warping field computed between two SMPLX bodies; this procedure follows Delta~\cite{feng2023learning}.

\begin{figure*}[ht!]
\centering 
\includegraphics[width=0.4\linewidth]{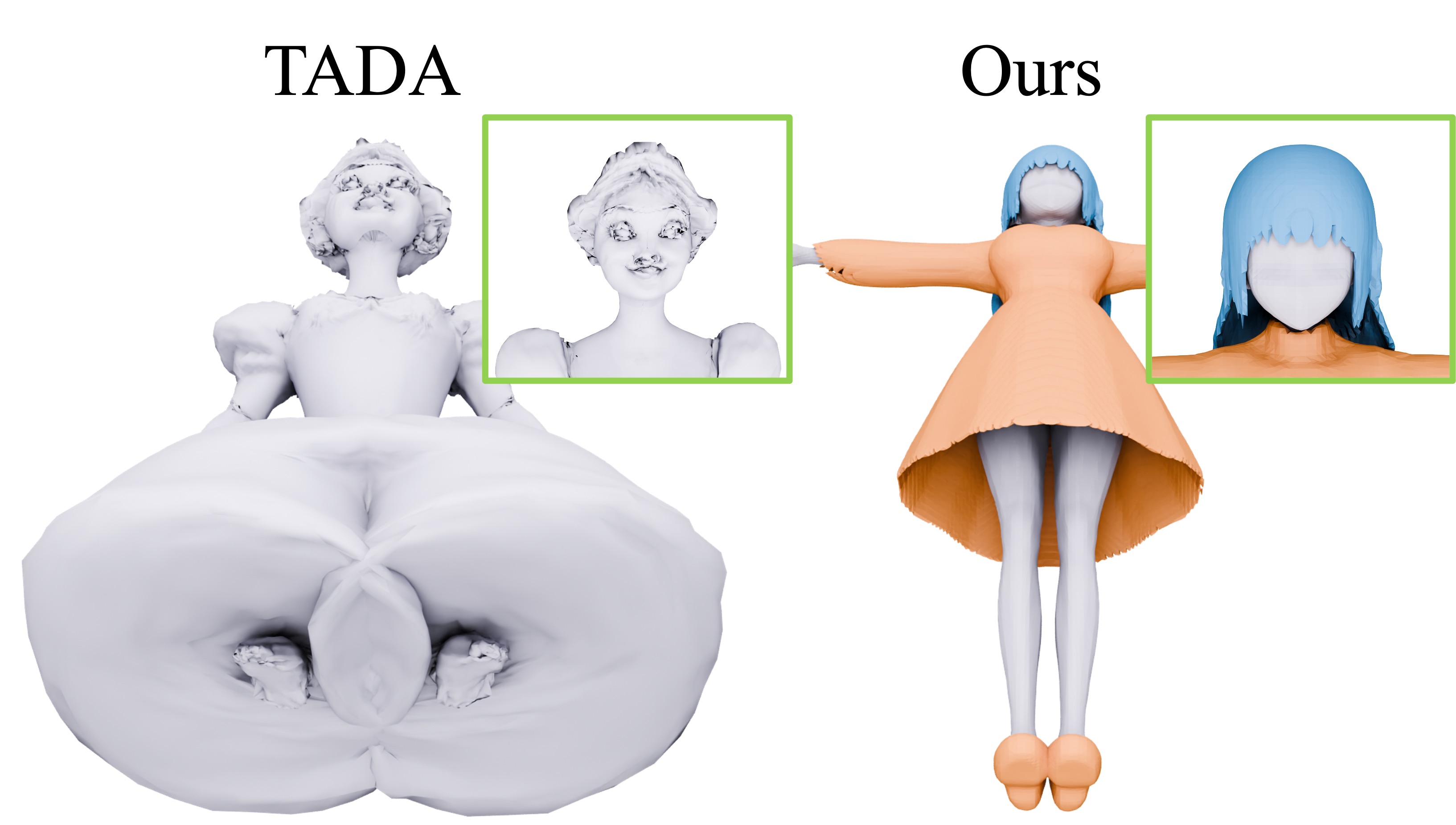}
  
\caption{ \textbf{Qualitative avatar generation results.} TADA~\cite{liao2024tada} generates defective and fragmented meshes when creating intricate hair and garments. Ours generates more smooth geometry and the generated shape is compositional.}
  \label{fig:avatar-gen-comparison}
\end{figure*}

We also compare with baseline TADA~\cite{liao2024tada}, a holistic avatar generation approach using SDS optimization.
Fig.~\ref{fig:avatar-gen-comparison} presents the qualitative avatar generation results.
Overall our model produces more natural geometry than TADA.
As TADA is designed to generate the entire body, it would result in defective and fragmented meshes when creating intricate hair and garments.
Tab.~\ref{tab:quan-avatar} numerically demonstrates the advantages of our method in terms of generation speed over TADA.

\subsection{Ablation Study} \label{sec:ablation}

\begin{figure}[ht!]
  \centering
    \includegraphics[width=0.8\linewidth]{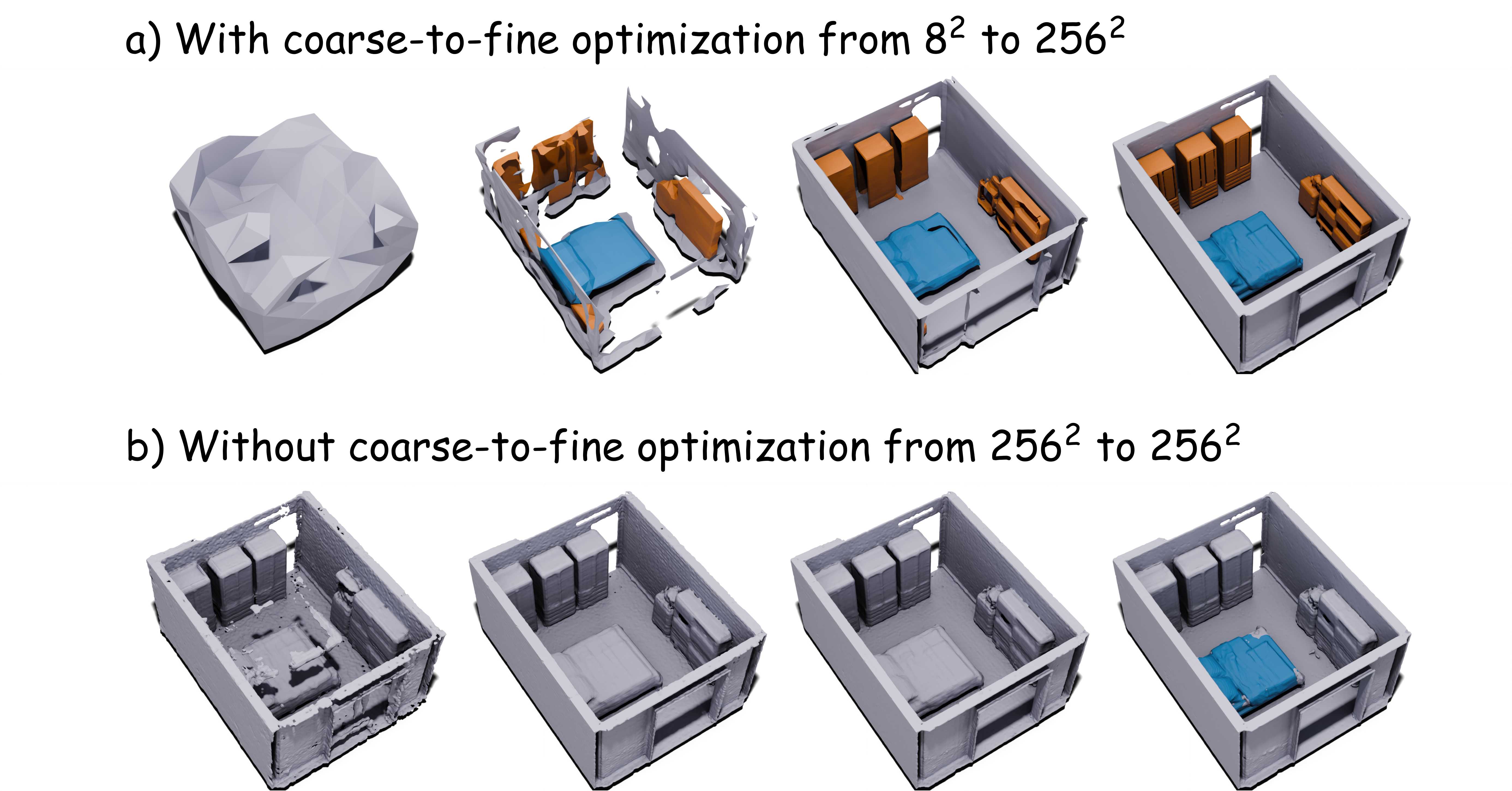}
  \caption{Visualization of the whole room (\textcolor[RGB]{149,148,160}{wall}, \textcolor[RGB]{85,139,184}{bed}, \textcolor[RGB]{206,143,95}{cabinet}) during tri-plane fitting with or without coarse-to-fine optimization.}
  \label{fig:change}
\end{figure}

\smallskip
\noindent
\textbf{Is coarse-to-fine optimization necessary?}
Fig.~\ref{fig:change} illustrates the reconstructed rooms during tri-plane fitting, with and without coarse-to-fine optimization. 
Optimizing tri-planes directly at high resolution fails to correct misclassified meshes that appear in the early training stage due to the gradient locality issue. 
By starting at a low resolution, the room's approximate semantics and shape can be outlined and subsequently refined smoothly at higher resolutions. 
Tab.~\ref{tab:tf} quantitatively proves the necessity.
Additionally, coarse-to-fine optimization reduces the fitting time from 380s to 279s over 500 iterations.

\begin{figure*}[htb]
  \centering
  \includegraphics[width=1\linewidth]{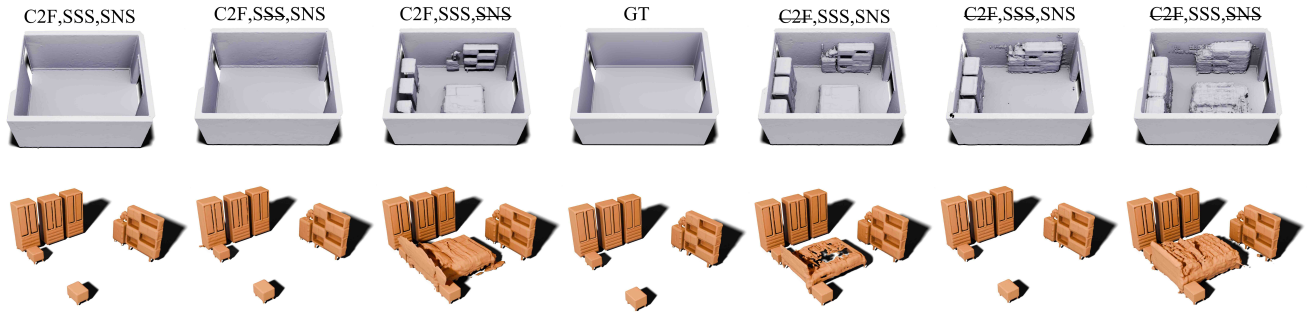}
  \caption{\textbf{Ablation studies on tri-plane fitting}. 
  As the generated beds are nearly the same in all experiments, we only show the results of \textcolor[RGB]{149,148,160}{wall} and \textcolor[RGB]{206,143,95}{cabinet}.}
  \label{fig:tf}
\end{figure*}

\smallskip
\noindent
\textbf{Ablation of tri-plane fitting.}\label{sec:abl-tf}
Tab.~\ref{tab:tf} and Fig.~\ref{fig:tf} show an ablation study for tri-plane fitting with different point sampling and optimization strategies.
We control the following 3 configurations with different combinations:
\textbf{C2F} indicates using \underline{c}oarse-\underline{t}o-\underline{f}ine optimization;
\textbf{SSS} indicates using \underline{s}emantic-aware \underline{s}ampling \underline{s}trategy that samples points uniformly for each category ($N_1\approx N_2\approx N_3$) rather than uniformly on the entire room ($N_1\gg N_2, N_1\gg N_3$);
\textbf{SNS} indicates \underline{s}ampling $\textbf{P}_{sdf}$ on both $[-1,1]^3$ cubic space and space \underline{n}ear \underline{s}urface rather than only on $[-1,1]^3$.
The \{C2F,SSS,\sout{SNS}\} combination demonstrates that coarse-to-fine optimization alone is insufficient to resolve semantic coupling, i.e., a bed may appear within a wall. 
Of the four terms in $\mathcal{L}_{tri}$, only $\mathcal{L}_{sdf}$ contributes to semantic isolation, which leaves room for the risk of semantic coupling. 
Since semantic coupling often manifests as the reflection of one category's surface in another, the comparison between \{C2F,SSS,SNS\} and \{C2F,SSS,\sout{SNS}\} proves that the additional ground truth SDF values near objects can alleviate this issue. 
\{C2F,\sout{SSS},SNS\} slightly outperforms \{C2F,SSS,SNS\} due to more sampled points on the wall under fitting task.
We will highlight the importance of SSS in VAE training later.

\begin{table}
    \caption{
    \textbf{Quantitative avatar generation results. }``Time'' records the duration of generating one avatar.}
    \label{tab:quan-avatar}
    \centering
    \scalebox{0.75}{
        \begin{tabular}{c|c|c}
        \toprule
        Method & TADA~\cite{liao2024tada} & Frankenstein (Ours)\\
        \midrule
        Time(min) $\downarrow$ & 271.6 & 0.5 \\
        Type & Holistic & Compositional \\
        \bottomrule
        \end{tabular}
    }
\end{table}

\begin{table}
    \caption{\textbf{Ablation studies on tri-plane fitting}. Chamfer Distance (CDs) are computed for each semantic group. $\mathcal{L}_{eik}$ and $\mathcal{L}_{sdf}$ are scaled by $10^{-2}$, and CDs are scaled by $10^{-3}$.} 
    \label{tab:tf}
    \centering
    \scalebox{0.8}{
        \begin{tabular}{ccc|cc|ccc}
        \toprule
        C2F & SSS & SNS & $\mathcal{L}_{eik}$ & $\mathcal{L}_{sdf}$ & CD$_{wall}$ & CD$_{bed}$ & CD$_{cabinet}$ \\
        \midrule
        \checkmark & \checkmark & \checkmark & 4.33 & 1.01 & 0.94 & \textbf{0.11} & \textbf{0.26}\\
         & \checkmark & \checkmark & 14.44 & 3.79 & 2.00 & 0.14 & 6.81\\
        \checkmark & & \checkmark & \textbf{2.32} & \textbf{0.31} & \textbf{0.89} & \textbf{0.11} & \textbf{0.26}\\
         & & \checkmark & 9.50 & 2.09 & 1.58 & 0.14 & 0.27\\
        \checkmark & \checkmark & & 6.20 & 0.55 & 1.67 & \textbf{0.11} & 8.74\\
         & \checkmark & & 9.17 & 1.28 & 2.25 & 0.13 & 8.62\\
        \bottomrule
        \end{tabular}
    }
\end{table}

\smallskip
\noindent
\textbf{Dimensions of the latent tri-plane.}
Tab.~\ref{tab:vae} presents the quantitative results of varying hyper-parameters on a small dataset (720 training rooms and 80 testing rooms) after 5k iterations. 
We explore different latent resolutions of $r=20,40,80$ and channel dimensions of $c=1,4,8$. 
The  latent's dimension greatly affects the generation results: a higher latent resolution leads to improved performance. 
While resolution with $80\times80$ imposes a significant computational overhead,  we regard $40\times40$ as a good trade-off between efficiency and quality. 
In the experiments on channel dimensions, we find that increasing number of  channels lead to more complete wall shape. 
To balance efficiency and the geometry quality, we set $c=4$.

\begin{table}
    \caption{
    \textbf{Ablation Studies on latent Tri-plane dimension}. Chamfer Distance (CD) is computed for wall/bed/cabinet and scaled by $10^{-3}$.}
    \label{tab:vae}
    \centering
    \scalebox{0.65}{
        \begin{tabular}{l|c|c|c|c|c}
        \toprule
        Latent Dim. & (4,20,20) & (1,40,40) & (4,40,40) & (8,40,40) & (4,80,80)\\
        \midrule
        CD(train)$\downarrow$ & 5.31/0.96/1.20 & 3.82/0.72/1.36 & 3.15/\textbf{0.52}/\textbf{0.94} & \textbf{1.81}/0.58/1.33  & \textbf{1.45}/\textbf{0.23}/\textbf{0.40} \\
        CD(test)$\downarrow$ & 6.04/1.52/2.06 & 3.72/1.54/4.00 & 4.02/\textbf{1.24}/\textbf{1.83} & \textbf{3.08}/1.32/2.09 & \textbf{2.30}/\textbf{0.83}/\textbf{1.00}\\
        \midrule
        Speed(s/iter)$\uparrow$ & 18 & 21 & 21 & 22 & 34 \\
        \bottomrule
        \end{tabular}
    }
\end{table}

\begin{figure*}[htb]
  \centering
  \includegraphics[width=0.5\linewidth]{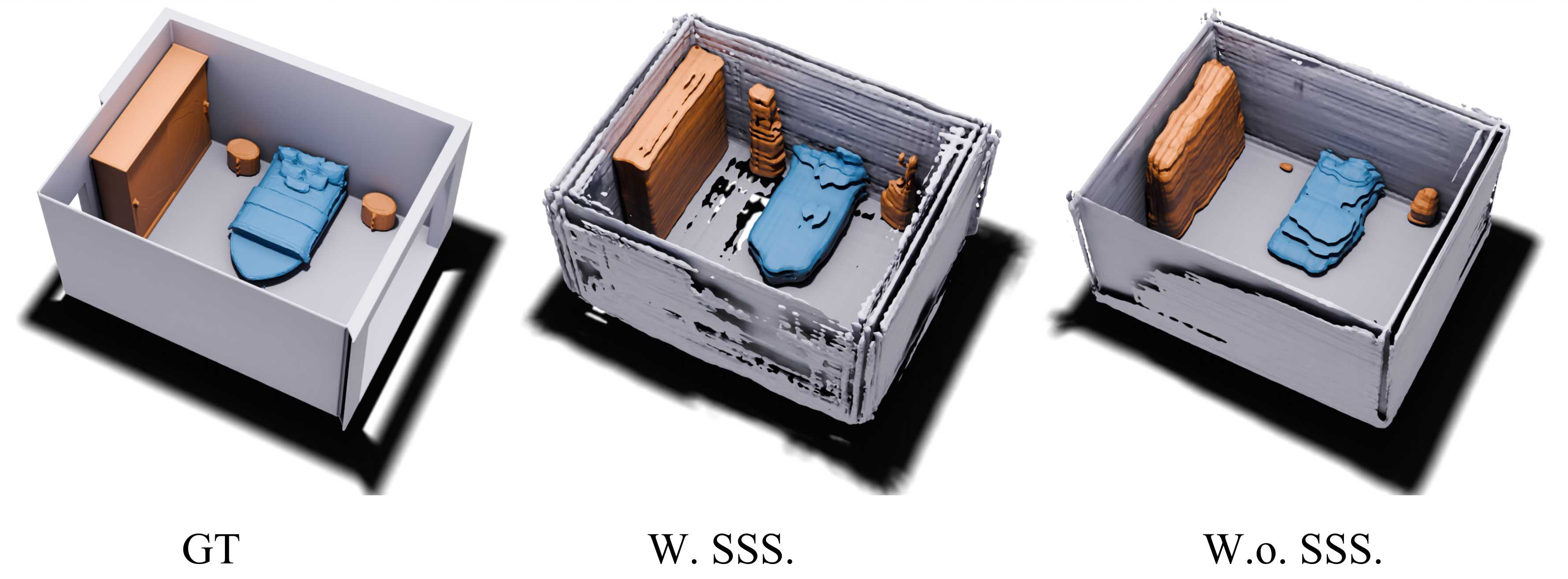}
\caption{Qualitative results on the semantic-aware sampling strategy (SSS).}
  \label{fig:SSS}
\end{figure*}

\smallskip
\noindent
\textbf{Is the semantic-aware sampling strategy necessary?} 
Fig.~\ref{fig:SSS} and Tab.~\ref{tab:SSS} demonstrate that the model has difficulty recovering the details of furniture without semantic-aware sampling strategy (SSS). 
The model is trained on a small dataset (720 training rooms and 80 testing rooms) for 5k iterations. 
Since walls dominate in terms of surface area compared to furniture, when we uniformly sample points across the entire mesh, most points are from walls, resulting in incomplete furniture (see Fig.~\ref{fig:SSS}). 
To address this, we sample $N_1:N_2:N_3=4:1:1$ to achieve a 
relatively balanced sampling across classes and calculate the loss function separately for each class. 
The results exhibit improved furniture details but rougher walls (see Fig.~\ref{fig:SSS}). 
We opt to apply SSS to enhance the details of the furniture.

\begin{table}
    \caption{\textbf{Quantitative results on the semantic-aware sampling strategy (SSS)}. CDs are computed for wall/bed/cabinet. The metrics are scaled by $10^{-3}$.
    } 
    \label{tab:SSS}
    \centering
    \scalebox{0.9}{
        \begin{tabular}{c|c|c}
        \toprule
         Model & CD(train)$\downarrow$ & CD(test)$\downarrow$\\
         \midrule
         w. SSS   & 3.15/\textbf{0.52}/\textbf{0.94}  & 4.02/\textbf{1.24}/\textbf{1.83} \\
         w.o. SSS & \textbf{2.78}/1.03/14.48 & \textbf{3.23}/1.83/8.59 \\
        \bottomrule
        \end{tabular}
    }
\end{table}

\begin{figure*}[htb]
  \centering
  \includegraphics[width=0.6\linewidth]{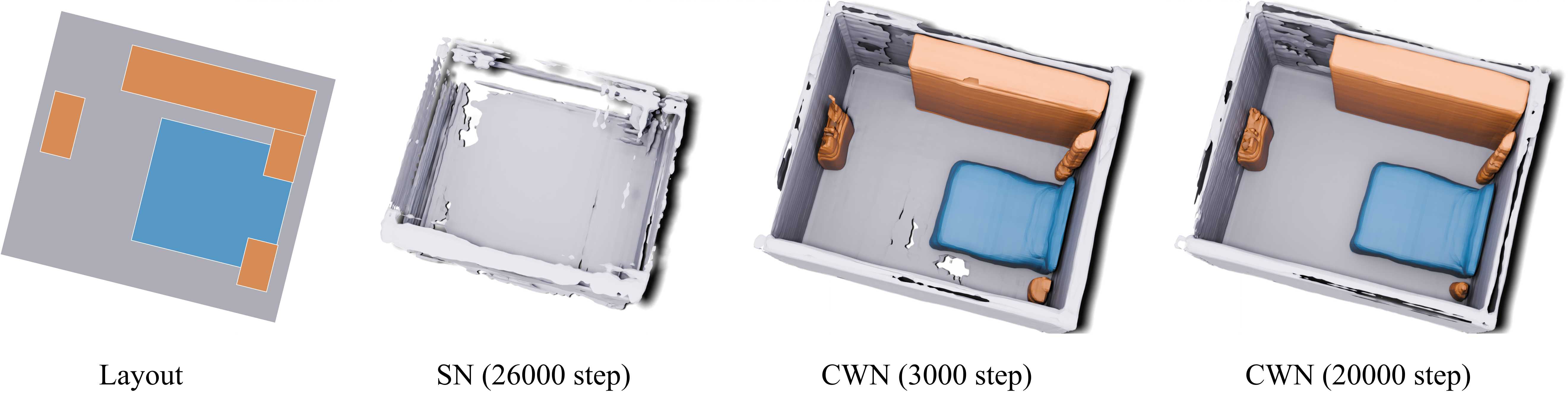}
\caption{Scale normalization (SN) v.s. channel-wise normalization (CWN).}
  \label{fig:nor}
\end{figure*}

\smallskip
\noindent
\textbf{How does the normalization effect the diffusion model?}
Scale normalization maintains the standard deviation of the data close to 1.0 by multiplying a scale coefficient~\cite{rombach2022high}. 
However, this approach is not effective for our task as it only generates walls of poor quality. 
We hypothesize that the wall feature dominates most of the $C$ channels, and other semantics may be overlooked if multiplied by a relatively large coefficient. 
With channel-wise normalization, all semantics can be effectively generated (see Fig.~\ref{fig:nor}). The model is trained on a small dataset ($\approx$ 2k latent tri-planes).

\smallskip
\noindent
\textbf{Is Frankenstein generalizable?}
Fig.~\ref{fig:diff} presents room generation results conditioned on various layout maps. 
Our model demonstrates the ability to adhere to a conditional layout while maintaining generation capacity when altering the layouts configuration.
We randomly select three layout maps from the training dataset and translate one of the objects along the wall, resulting in new layouts that remain reasonable. 
It is important to note that these new layouts do not exist in the training set. 
Conditioned on these new layouts, Frankenstein successfully generates corresponding 3D rooms that are reasonable. 
This showcases the model's generalization capacity.

\begin{figure*}[htb]
  \centering
    \includegraphics[width=0.5\linewidth]{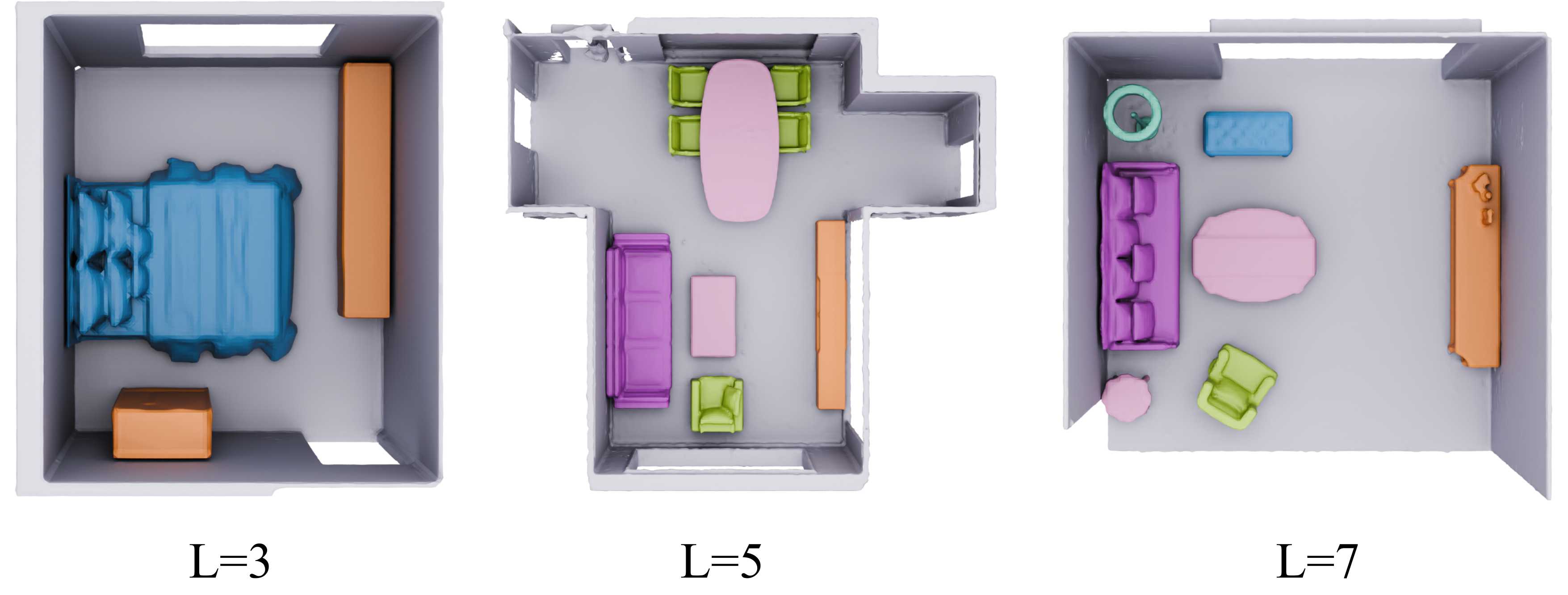}
\caption{\textbf{Tri-plane fitting results under different number of classes.} Each color represents an SDF field sharing identical semantic information.}
  \label{fig:multiclass}
\end{figure*}

\begin{figure*}[htb]
  \centering
    \includegraphics[width=0.8\linewidth]{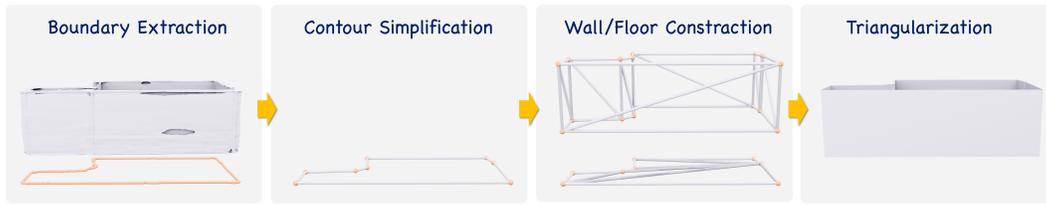}
\caption{\textbf{Wall remesh algorithm.} 1) 2D boundary of the generated wall is extracted using~\cite{van2014scikit}; 2) The boundary is simplified into few points using~\cite{ramer1972iterative,douglas1973algorithms}; 3) The wall and floor are constracted by connecting points in sequence and~\cite{eberly2008triangulation}.}
  \label{fig:remesh}
\end{figure*}

\begin{figure*}[htb]
  \centering
    \includegraphics[width=1\linewidth]{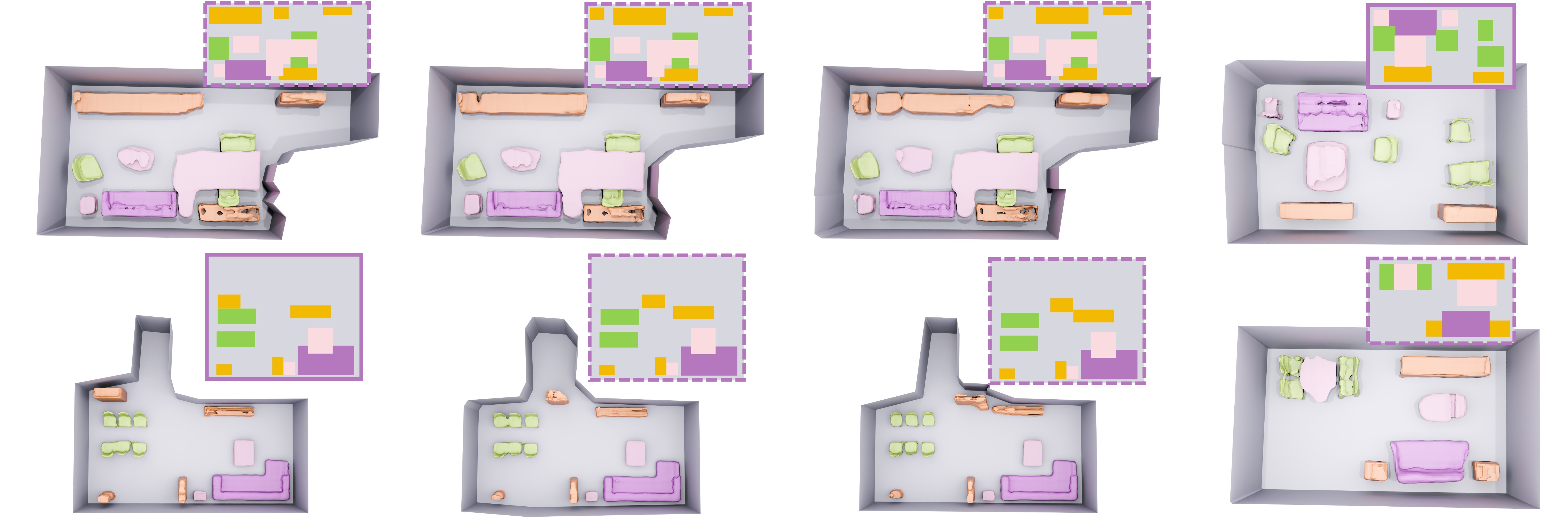}
\caption{\textbf{Generated rooms with more objects ($L=5$)}. The layout maps in solid line are from the training dataset.}
  \label{fig:multiclass-gen}
\end{figure*}

\smallskip
\noindent
\textbf{Capacity of Tri-plane for modeling semantic classes.} 
Fig.~\ref{fig:multiclass} demonstrates that one single tri-plane is capable to model multi-class scenes with $L$ set up to seven.
Fig.~\ref{fig:multiclass-gen} displays qualitative results for more complex scenes (L=5). 
The walls have been remeshed according to the outline of the generated mesh (see Fig.~\ref{fig:remesh}).

\smallskip
\noindent
\textbf{Number of training samples for MLP fitting in stage 1.} 
We first train the shared MLP in stage 1 (c.f. Section.~\ref{sec:stage1}) with 1, 10, 100 samples, respectively. Then the MLPs are used to fit another 100 rooms for testing. The evaluation is listed in Tab~\ref{tab:smlp}.
One sample is insufficient to learn a robust MLP, while 100 samples have no obvious improvement compared to using 10 samples.
Considering time efficiency, we choose 10 samples to train the shared MLP.

\begin{table}[h!]
    \caption{ \textbf{Ablation study on the shared MLP}. The values of CD are scaled by $10^{-3}$. The unit of time is minute.}
    \label{tab:smlp}
    \centering
    \scalebox{0.8}{
        \begin{tabular}{l|c|c|c|c}
        \toprule
         \# Training Samples & $CD_{mesh}\downarrow$ & $CD_{bed}\downarrow$ & $CD_{Cabinet}\downarrow$ & Time$\downarrow$\\
         \midrule
         1   & 1.99 & 0.19 & 0.19 & 1.48\\
         10  & 0.79 & 0.11 & 0.11 & 4.38\\
         100 & 0.79 & 0.11 & 0.11 & 32.77\\
        \bottomrule
        \end{tabular}
    }
\end{table}

\section{Conclusion}

In summary, we develop Frankenstein, an innovative tri-plane diffusion-based method that is capable of generating semantic-compositional 3D scenes.
We extend the tri-plane tensor factorization to represent compositional shapes by decoding multiple SDFs from a single tri-plane, each of which represents the shape of a distinct semantic class.
This novel representation permits the concurrent modelling of multiple comprehensive shapes within a single tri-plane tensor, ensuring distinct scene separation and a diverse range of generated shapes.
The generated scenes offer a range of customized controls, including but not limited to, component-wise texturing, rearrangement of room objects, and the re-targeting of avatar clothing and hair.

\begin{figure*}[htb]
  \centering
    \includegraphics[width=0.8\linewidth]{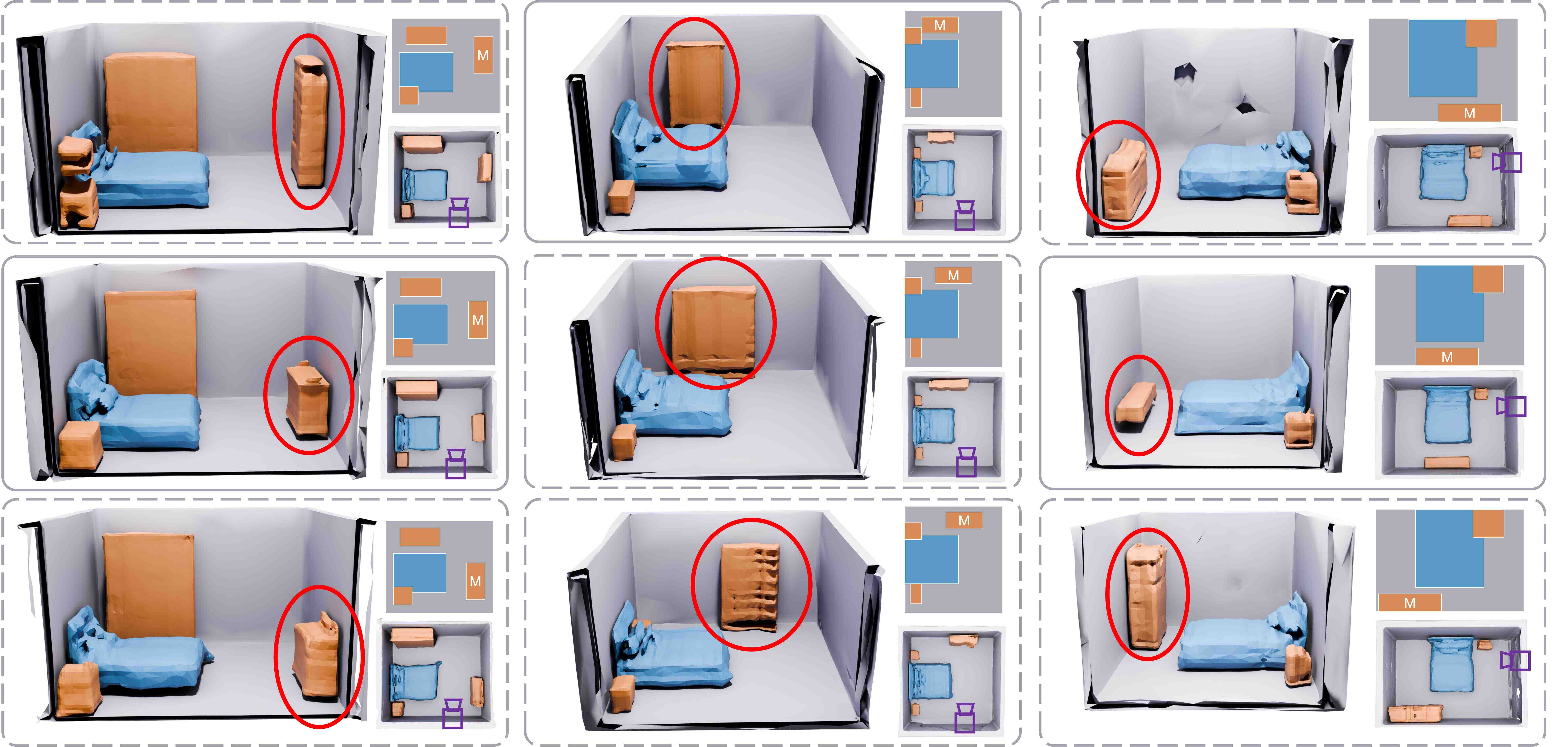}
\caption{
\textbf{Qualitative results of layout-conditioned room generation.} 
The layout maps in solid line are from the training dataset, and the other two layout maps in the same column are generated by translating one of the furniture (marked as ``M'').}
  \label{fig:diff}
\end{figure*}

\smallskip
\noindent
\textbf{Limitations.}
The current implementation of Frankenstein faces several limitations.
1) Given that our method utilizes only a single tri-plane to model the entire scene, the details are limited by the resolution. 
One potential solution could involve combining our approach with~\cite{wu2024blockfusion}, which divides large scenes into multiple smaller blocks.
2) The training process for the VAE is slow ($\approx$ a week), incorporating more efficient backbone is a future direction.
3) While our method demonstrates generalizability in layout conditions, it still exhibits some flaws (e.g. holes in the wall in Fig.~\ref{fig:diff}) with layouts beyond the dataset.
We believe that these issues can be mitigated through the use of additional training data.

\begin{acks}
This work was supported in part by NSFC (62322113, 62376156) and Shanghai Municipal Science and Technology Major Project (2021SHZDZX0102).
%
%
%
%
\end{acks}

\bibliographystyle{ACM-Reference-Format}
\bibliography{sample-bibliography}



